\documentclass[10pt,twocolumn,letterpaper]{article}

\usepackage{iccv}
\usepackage{times}
\usepackage{epsfig}
\usepackage{graphicx}
\usepackage{amsmath}
\usepackage{amssymb}

\usepackage[breaklinks=true,bookmarks=false]{hyperref}

\iccvfinalcopy 

\def\iccvPaperID{****} 

\ificcvfinal\fi

\begin{document}

\title{Visual ChatGPT: Talking, Drawing and Editing with Visual Foundation Models}

\author{Chenfei Wu \quad Shengming Yin \quad Weizhen Qi \quad Xiaodong Wang \quad Zecheng Tang \quad Nan Duan\thanks{Corresponding author.}
\\
Microsoft Research Asia \\
\{chewu, v-sheyin, t-weizhenqi, v-xiaodwang, v-zetang, nanduan\}@microsoft.com}

\maketitle

\begin{abstract}

ChatGPT is attracting a cross-field interest as it provides a language interface with remarkable conversational competency and reasoning capabilities across many domains. However, since ChatGPT is trained with languages, it is currently not capable of processing or generating images from the visual world. At the same time, Visual Foundation Models, such as Visual Transformers or Stable Diffusion, although showing great visual understanding and generation capabilities, they are only experts on specific tasks with one-round fixed inputs and outputs. To this end, We build a system called \textbf{Visual ChatGPT}, incorporating different Visual Foundation Models, to enable the user to interact with ChatGPT by 1) sending and receiving not only languages but also images 2) providing complex visual questions or visual editing instructions that require the collaboration of multiple AI models with multi-steps. 3) providing feedback and asking for corrected results. We design a series of prompts to inject the visual model information into ChatGPT, considering models of multiple inputs/outputs and models that require visual feedback. Experiments show that Visual ChatGPT opens the door to investigating the visual roles of ChatGPT with the help of Visual Foundation Models. Our system is publicly available at \url{https://github.com/microsoft/visual-chatgpt}.

\end{abstract}

\section{Introduction}

In recent years, the development of Large language models (LLMs) has shown incredible progress, such as T5~\cite{raffel2020exploring}, BLOOM~\cite{scao2022bloom}, and GPT-3~\cite{brown2020language}. One of the most significant breakthroughs is ChatGPT, which is built upon InstructGPT~\cite{ouyang2022training}, specifically trained to interact with users in a genuinely conversational manner, thus allowing it to maintain the context of the current conversation, handle follow-up questions, and correct answer produced by itself.  

Although powerful, ChatGPT is limited in its ability to process visual information since it is trained with a single language modality, while Visual Foundation Models~(VFMs) have shown tremendous potential in computer vision, with their ability to understand and generate complex images. For instance, BLIP Model~\cite{li2023blip} is an expert in understanding and providing the description of an image. Stable Diffusion~\cite{rombach2022high} is an expert in synthesizing an image based on text prompts. However, suffering from the task specification nature, the demanding and fixed input-output formats make the VFMs less flexible than conversational language models in human-machine interaction.

\begin{figure}
\centering
\includegraphics[width=0.45\textwidth]{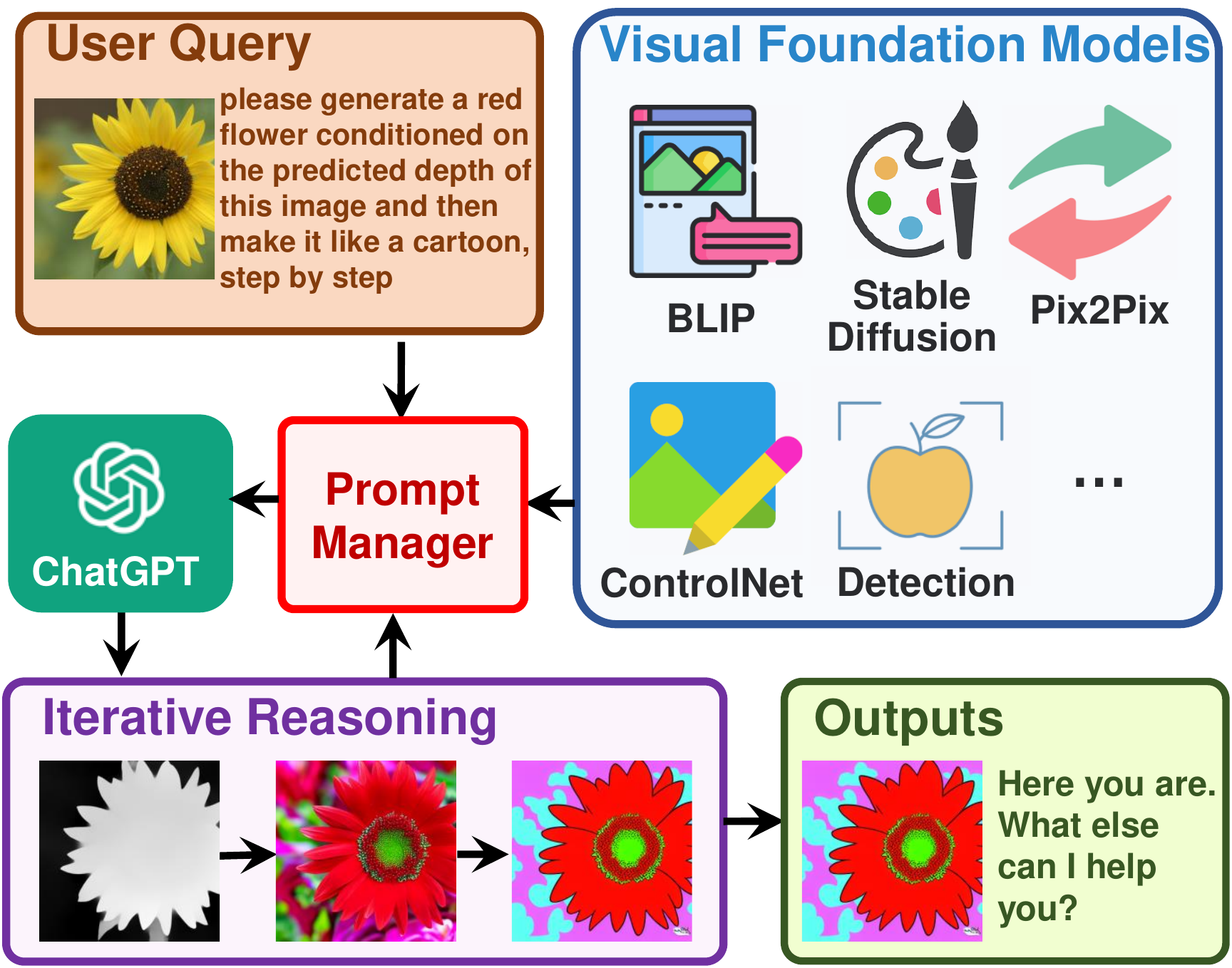} 
\caption{Architecture of Visual ChatGPT.} 
\label{fig:arch}
\vspace{-8mm}
\end{figure}

Could we build a ChatGPT-like system that also supports image understanding and generation? One intuitive idea is to train a multi-modal conversational model. However, building such a system would consume a large amount of data and computational resources. Besides, another challenge comes that what if we want to incorporate modalities beyond languages and images, like videos or voices? Would it be necessary to train a totally new multi-modality model every time when it comes to new modalities or functions?

We answer the above questions by proposing a system named \textbf{Visual ChatGPT}. Instead of training a new multi-modal ChatGPT from scratch, we build Visual ChatGPT directly based on ChatGPT and incorporate a variety of VFMs. To bridge the gap between ChatGPT and these VFMs, we propose a Prompt Manager which supports the following functions: 1) explicitly tells ChatGPT 
the capability of each VFM and specifies the input-output formats; 2) converts different visual information, for instance, png images, the depth images and mask matrix, to language format to help ChatGPT understand; 3) handles the histories, priorities, and conflicts of different Visual Foundation Models. With the help of the Prompt Manager, ChatGPT can leverage these VFMs and receives their feedback in an iterative manner until it meets the requirements of users or reaches the ending condition. 



As shown in Fig.~\ref{fig:arch}, a user uploads an image of a yellow flower and enters a complex language instruction ``please generate a red flower conditioned on the predicted depth of this image and then make it like a cartoon, step by step''. With the help of Prompt Manager, Visual ChatGPT starts a chain of execution of related Visual Foundation Models. In this case, it first applies the depth estimation model to detect the depth information, then utilizes the depth-to-image model to generate a figure of a red flower with the depth information, and finally leverages the style transfer VFM based on the Stable Diffusion model to change the style of this image into a cartoon. During the above pipeline, Prompt Manager serves as a dispatcher for ChatGPT by providing the type of visual formats and recording the process of information transformation. Finally, when Visual ChatGPT obtains the hints of ``cartoon'' from Prompt Manager, it will end the execution pipeline and show the final result.

In summary, our contributions are as follows:

\begin{itemize}
\item{We propose Visual ChatGPT, which opens the door of combining ChatGPT and Visual Foundation Models and enables ChatGPT to handle complex visual tasks;}
\item{We design a  Prompt Manager, in which we involve \textbf{22} different VFMs and define the internal correlation among them for better interaction and combination;}
\item{Massive zero-shot experiments are conducted and abundant cases are shown to verify the understanding and generation ability of Visual ChatGPT.}
\end{itemize}

\begin{figure*}
\centering
\includegraphics[width=\textwidth]{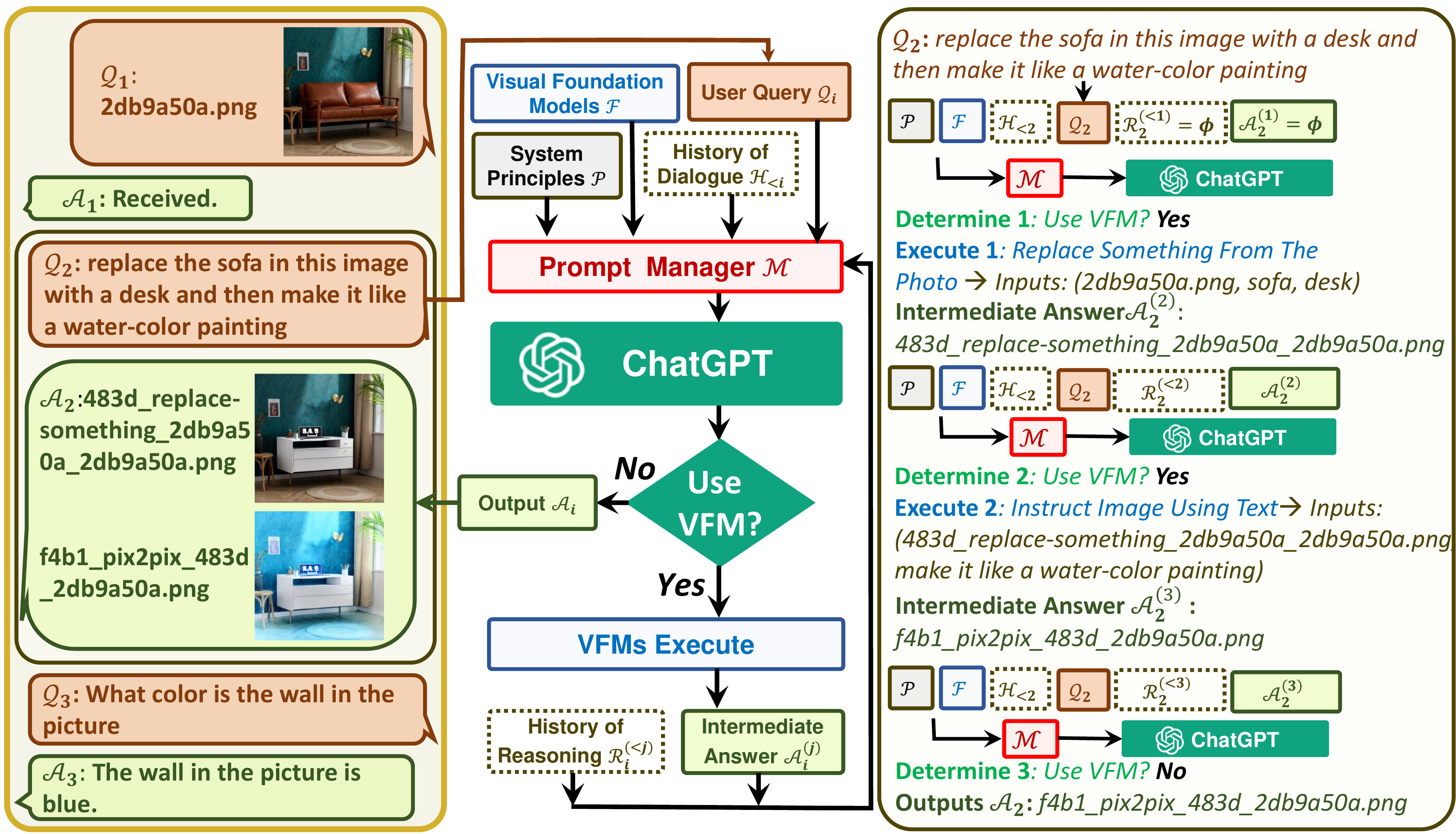} 
\caption{Overview of Visual ChatGPT. The left side shows a three-round dialogue, The middle side shows the flowchart of how Visual ChatGPT iteratively invokes Visual Foundation Models and provide answers. The right side shows the detailed process of the second QA.} 

\label{fig:chat_framework}
\end{figure*}

\vspace{-6mm}

\section{Related Works}
\vspace{-1mm}
\subsection{Natural Language and Vision}
\vspace{-1mm}
Surrounded by various modalities~(sound, vision, video, \textit{etc}), language and vision are the two main mediums transmitting information in our life. There is a natural link between the natural language and visions, and most questions require joint modeling of both two streams to produce the satisfied results~\cite{gan2017semantic,lisupervision,zellers2019recognition}, e.g., visual question answering~(VQA)~\cite{antol2015vqa} takes an image and one corresponding question as input and requires to generate an answer according to the information in the given image. Owing to the success of large language models~(LLMs) like InstructGPT~\cite{ouyang2022training}, one can easily interact with the model or obtain feedback in the natural language format, but it is incapable for those LLMs to process the visual information. To fuse the vision processing ability into such LLMs, several challenges are lying ahead since it is hard to train either large language models or vision models, and the well-designed instructions~\cite{brooks2022instructpix2pix,zhang2021text,li2020manigan} and cumbersome conversions~\cite{radford2021learning,zhai2022lit} are required to connect different modalities.
Although several works have explored leveraging the pre-trained LLMs to improve the performance on the vision-language~(VL) tasks, those methods supported several specific VL tasks (from language to version or from version to language) and required labeled data for training~\cite{tsimpoukelli2021multimodal,alayracflamingo,li2023blip}.

\vspace{-1mm}
\subsection{Pre-trained Models for VL tasks}
\vspace{-1mm}
To better extract visual features, frozen pre-trained image encoders are adopted in the early works~\cite{chen2020uniter,li2020oscar,zhang2021vinvl}, and recent LiT~\cite{zhai2022lit} apply the CLIP pre-training~\cite{radford2021learning} with frozen ViT model~\cite{zhai2022scaling}.
From another perspective, exploiting the knowledge from LLMs also counts. 
Following the instruction of Transformer~\cite{vaswani2017attention}, pre-trained LLMs demonstrate a powerful text understanding and generation capability~\cite{radford2019language,kenton2019bert,stiennon2020learning,brown2020language}, and such breakthroughs also benefit the VL modelling~\cite{fu2021violet,gan2020large,bao2021vlmo,zellers2022merlot}, where these works add an extra adapter modules~\cite{houlsby2019parameter} in the pre-trained LLMs to align visual features to the text space.
With the increased number of model parameters, it is hard to train those pre-trained LLMs, thus more efforts have been paid to directly leverage the off-the-shelf frozen pre-trained LLMs for VL tasks~\cite{eichenberg2021magma,tsimpoukelli2021multimodal,chen2022visualgpt,yang2022empirical,zeng2022socratic}.
\vspace{-1mm}
\subsection{Guidance of Pre-trained LLMs for VL tasks}
\vspace{-1mm}
To deal with complex tasks, e.g., commonsense reasoning~\cite{davis2015commonsense}, Chain-of-Thought~(CoT) is proposed to elicit the multi-step reasoning abilities of LLMs~\cite{weichain2022}. More concretely, CoT asks the LLMs to generate the intermediate answers for the final results. Existing study~\cite{zhang2023multimodal} have divided such a technique into two categories: Few-Shot-CoT~\cite{zhang2022automatic} and Zero-Shot-CoT~\cite{kojimalarge2022}.
For the few-shot setting, the LLMs perform CoT reasoning with several demonstrations~\cite{zhou2022least, wang2022self}, and it turns out that the LLMs can acquire better abilities to solve complex problems. Further, recent studies~\cite{kojimalarge2022, zelikmanstar} have shown that LLMs can be self-improved by leveraging self-generated rationales under the zero-shot setting. 
The above studies mainly focus on a single modality, i.e., language. Recently, Multimodal-CoT~\cite{zhang2023multimodal} is proposed to incorporate language and vision modalities into a two-stage framework that separates rationale generation and answer inference.
However, such a method merely shows superiority under specific scenarios, i.e., ScienceQA benchmark~\cite{lulearn2022}.
In a nutshell, our work extends the potentiality of CoT to massive tasks, including but not limited to text-to-image generation~\cite{lin2014microsoft}, image-to-image translation~\cite{isola2017image}, image-to-text generation~\cite{vinyals2015show}, \textit{etc.}

\begin{figure*}
\centering
\includegraphics[width=\textwidth]{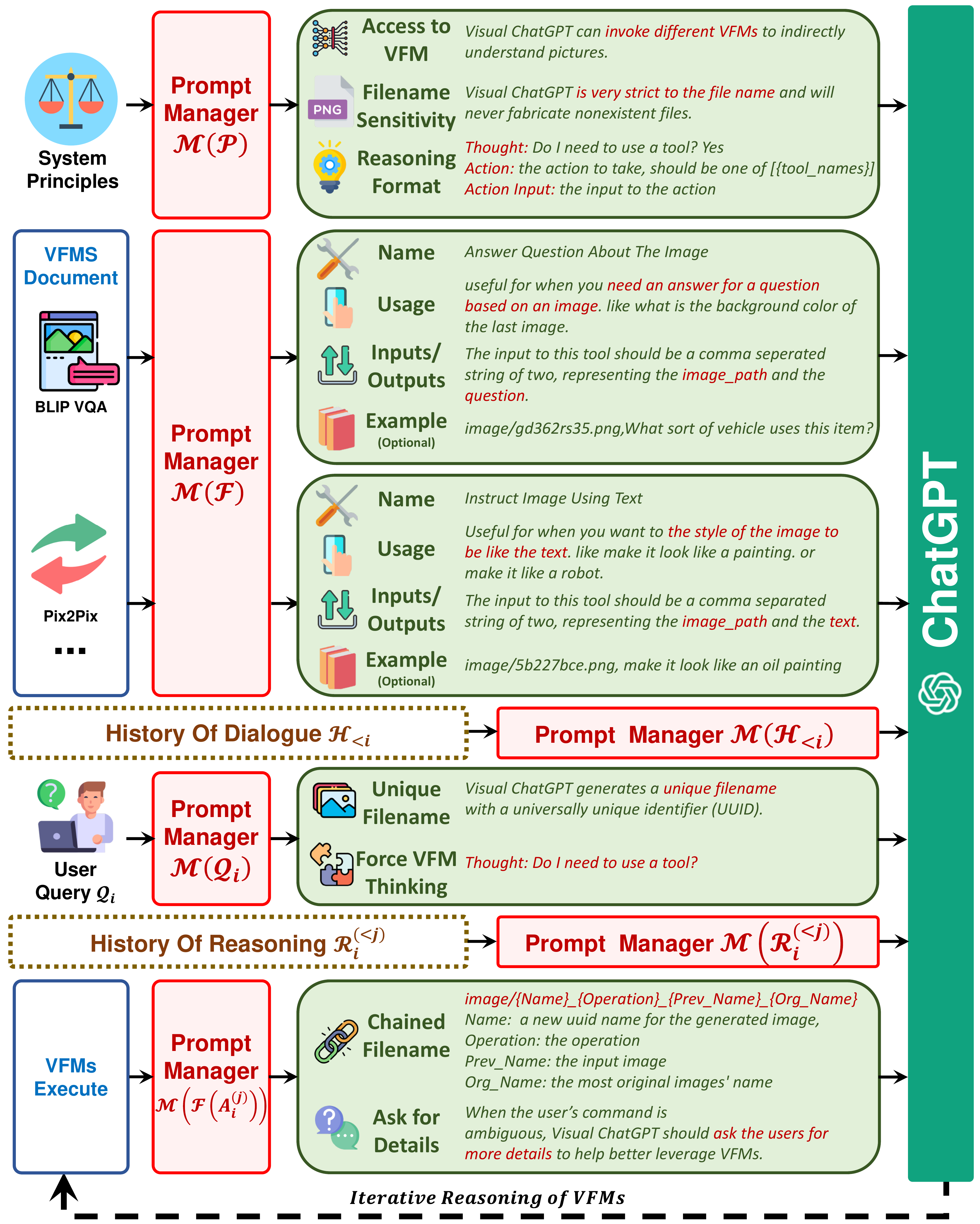} 
\caption{Overview of Prompt Manager. It coverts all non-language signals into language so that ChatGPT can understand.} 
\label{fig:M(F)}
\end{figure*}
\section{Visual ChatGPT}

Let $\mathcal{S} = \{(\mathcal{Q}_1,\mathcal{A}_1), (\mathcal{Q}_2,\mathcal{A}_2), ...,(\mathcal{Q}_N,\mathcal{A}_N)\}$ be a dialogue system with $N$ question-answer pairs. To get the response $\mathcal{A}_i$ from the $i$-th round of conversation, a series of VFMs and intermediate outputs $\mathcal{A}_{i}^{(j)}$ from those models are involved, where $j$ denotes the output from the $j$-th VFM~($\mathcal{F}$) in $i$-th round.
More concretely, handling with Prompt Manager $\mathcal{M}$,  the format of $\mathcal{A}_{i}^{(j)}$ is constantly modified to meet the input format of each $\mathcal{F}$.
In the end, the system output $\mathcal{A}_{i}^{(j)}$ if it is denoted as the final response, and no more VFM is executed.
Eq.~(\ref{eq:a}) provides a formal definition of Visual ChatGPT:


\begin{align}\label{eq:a}
\mathcal{A}_{i}^{(j+1)} = ChatGPT(&\mathcal{M}(\mathcal{P}), \mathcal{M}(\mathcal{F}),\mathcal{M}(\mathcal{H}_{<i}),\mathcal{M}(\mathcal{Q}_i),  \nonumber \\ 
&\mathcal{M}(\mathcal{R}_{i}^{(<j)}), \mathcal{M}(\mathcal{F}(\mathcal{A}_{i}^{(j)})))
\end{align}




\textbf{-- System Principle $\mathcal{P}$}: System Principle provides basic rules for Visual ChatGPT, e.g., it should be sensitive to the image filenames, and should use VFMs to handle images instead of generating the results based on the chat history.

\textbf{-- Visual Foundation Model $\mathcal{F}$}: One core of Visual ChatGPT is the combination of various VFMs: $\mathcal{F}=\{f_1, f_2, ..., f_N\}$, where each foundation model $f_i$ contains a determined function with explicit inputs and outputs. 

\textbf{-- History of Dialogue $\mathcal{H}_{<i}$}: We define the dialogue history of $i$-th round of conversation as the string concatenation of previous question answer pairs, i.e, $\{(\mathcal{Q}_1,\mathcal{A}_1), (\mathcal{Q}_2,\mathcal{A}_2), \cdots,(\mathcal{Q}_{i-1},\mathcal{A}_{i-1})\}$. Besides, we truncate the dialogue history with a maximum length threshold to meet the input length of ChatGPT model.

\textbf{-- User query $\mathcal{Q}_{i}$}: In visual ChatGPT, query is a general term, since it can include both linguistic and visual queries. For instance, Fig.~\ref{fig:arch} shows an example of a query containing both the query text and the corresponding image.

\textbf{-- History of Reasoning $\mathcal{R}_{i}^{(<j)}$}: To solve a complex question, Visual ChatGPT may require the collaboration of multiple VFMs. For the $i$-th round of conversation, $\mathcal{R}_{i}^{(<j)}$ is all the previous reasoning histories from $j$ invoked VFMs.

\textbf{-- Intermediate Answer $\mathcal{A}^{(j)}$}: When handling a complex query, Visual ChatGPT will try to obtain the final answer step-by-step by invoking different VFMs logically, thus producing multiple intermediate answers. 

\textbf{-- Prompt Manager $\mathcal{M}$}: A prompt manager is designed to convert all the visual signals into language so that ChatGPT model can understand. In the following subsections, we focus on introducing how $\mathcal{M}$ manages above different parts: $\mathcal{P}, \mathcal{F}, \mathcal{Q}_i, \mathcal{F}(\mathcal{A}_{i}^{(j)})$.


\subsection{Prompt Managing of System Principles $\mathcal{M}(\mathcal{P})$} \label{sec.MP}
Visual ChatGPT is a system that integrates different VFMs to understand visual information and generation corresponding answers. To accomplish this, some system principles need to be customized, which are then transferred into prompts that ChatGPT can understand. These prompts serve several purposes, including:

\begin{itemize}
\item{\textbf{Role of Visual ChatGPT}} Visual ChatGPT is designed to assist with a range of text and visual-related tasks, such as VQA, image generation, and editing. 

\item{\textbf{VFMs Accessibility}} Visual ChatGPT has access to a list of VFMs to solve various VL tasks. The decision of which foundation model to use is entirely made by the ChatGPT model itself, thus making it easy to support new VFMs and VL tasks.

\item{\textbf{Filename Sensitivity}} Visual ChatGPT accesses image files according to the filename, and it is crucial to use precise filenames to avoid ambiguity since one round of conversation may contain multiple images and their different updated versions and the misuse of filenames will lead to the confusion about which image is currently being discussed. Therefore, Visual ChatGPT is designed to be strict about filename usage, ensuring that it retrieves and manipulates the correct image files. 

\item{\textbf{Chain-of-Thought}} As shown in Fig.~\ref{fig:arch}, to cope with one seemingly simple command may require multiple VFMs, e.g., the query of ``generate a red flower conditioned on the predicted depth of this image and then make it like a cartoon " requires depth estimation, depth-to-image and the style transfer VFMs. To tackle more challenging queries by decomposing them into subproblems, CoT is introduced in Visual ChatGPT to help decide, leverage and dispatch multiple VFMs.


\item{\textbf{Reasoning Format Strictness}} Visual ChatGPT must follow strict reasoning formats. Thus we parse the intermediate reasoning results with the elaborate regex matching algorithms, and construct the rational input format for ChatGPT model to help it determine the next execution, e.g., triggering a new VFM or returning the final response.

\item{\textbf{Reliability}} As a language model, Visual ChatGPT may fabricate fake image filenames or facts, which can make the system unreliable. To handle such issues, we design prompts that require Visual ChatGPT to be loyal to the output of the vision foundation models and not fabricate image content or filenames. Besides, the collaboration of multiple VFMs can increase system reliability, thus the prompt we construct will guide ChatGPT to leverage VFMs preferentially instead of generating results based on conversation history.

\end{itemize}




\subsection{Prompt Managing of Foundation Models $\mathcal{M}(\mathcal{F})$}

\begin{table}[]
\centering
\caption{Foundation models supported by Visual ChatGPT.}
\begin{tabular}{|ll|}
\hline
\multicolumn{2}{|l|}{Remove Objects from Image~\cite{cheng2021per,rombach2022high}  }          \\ \hline
\multicolumn{2}{|l|}{Replace Objects from Image~\cite{cheng2021per,rombach2022high}}         \\ \hline
\multicolumn{2}{|l|}{Change Image by the Text~\cite{rombach2022high}}            \\ \hline
\multicolumn{2}{|l|}{Image Question Answering~\cite{li2022blip}}              \\ \hline
\multicolumn{1}{|l|}{Image-to-Text~\cite{li2022blip}}      & Text-to-Image~\cite{rombach2022high}      \\ \hline
\multicolumn{1}{|l|}{Image-to-Edge~\cite{xu2017canny}}      & Edge-to-Image~\cite{zhang2023adding}      \\ \hline
\multicolumn{1}{|l|}{Image-to-Line~\cite{gu2022towards}}      & Line-to-Image~\cite{zhang2023adding}      \\ \hline
\multicolumn{1}{|l|}{Image-to-Hed~\cite{xie2015holistically}}       & Hed-to-Image~\cite{zhang2023adding}       \\ \hline
\multicolumn{1}{|l|}{Image-to-Seg~\cite{li2022uniformer}}       & Seg-to-Image~\cite{zhang2023adding}       \\ \hline
\multicolumn{1}{|l|}{Image-to-Depth~\cite{ranftl2020towards,ranftl2021vision}}     & Depth-to-Image~\cite{zhang2023adding}     \\ \hline
\multicolumn{1}{|l|}{Image-to-NormalMap~\cite{ranftl2020towards,ranftl2021vision}} & NormalMap-to-Image~\cite{zhang2023adding} \\ \hline
\multicolumn{1}{|l|}{Image-to-Sketch~\cite{xie2015holistically}}    & Sketch-to-Image~\cite{zhang2023adding}    \\ \hline
\multicolumn{1}{|l|}{Image-to-Pose~\cite{cao2017realtime}}      & Pose-to-Image~\cite{zhang2023adding}      \\ \hline
\end{tabular}

\label{tab:foundation}
\end{table}

Visual ChatGPT is equipped with multiple VFMs to handle various VL tasks. Since these different VFMs may share some similarities, e.g., the replacement of objects in the image can be regarded as generating a new image, and both Image-to-Text (I2T) task and Image Question Answering (VQA) task can be understood as giving the response according to the provided image, it is critical to distinguish among them. As shown in Fig.~\ref{fig:M(F)}, the Prompt Manager specifically defines the following aspects to help Visual ChatGPT accurately understand and handle the VL tasks:


\begin{itemize}
\item{\textbf{Name}} The name prompt provides an abstract of the overall function for each VFM, e.g., answer question about the image, and it not only helps Visual ChatGPT to understand the purpose of VFM in a concise manner but also severs as the entry to VFM.

\item{\textbf{Usage}} The usage prompt describes the specific scenario where the VFM should be used. For example, the Pix2Pix model~\cite{rombach2022high} is suitable for changing the style of an image. Providing this information helps Visual ChatGPT make informed decisions about which VFM to use for the particular task.

\item{\textbf{Inputs/Outputs}} The inputs and the outputs prompt outlines the format of inputs and outputs required by each VFM since the format can vary significantly and it is crucial to provide clear guideline for Visual ChatGPT to execute the VFMs correctly. 


\item{\textbf{Example(Optional)}}  The example prompt is optional, but it can be helpful for Visual ChatGPT to better understand how to use particular VFM under the specific input template and deal with more complex queries.




\end{itemize}


\subsection{Prompt Managing of User Querie $\mathcal{M}(\mathcal{Q}_i)$}

Visual ChatGPT supports a variety of user queries, including languages or images, simple or complex ones, and the reference of multiple images. Prompt Manager handles user queries in the following two aspects:

\begin{itemize}
\item{\textbf{Generate Unique Filename}}  Visual ChatGPT can handle two types of image-related queries: those that involve newly uploaded images and those that involve reference to existing images. For newly uploaded images, Visual ChatGPT generates a unique filename with a universally unique identifier~(UUID) and adds a prefix string "image" representing the relative directory, e.g., "image/\{uuid\}.png". Although the newly uploaded image will not be fed into ChatGPT, a fake dialogue history is generated with a question stating the image's filename and an answer indicating that the image has been received. This fake dialogue history assists in the following dialogues. For queries that involve reference to existing images, Visual ChatGPT ignores the filename check. This approach has been proven beneficial since ChatGPT has the ability to understand fuzzy matching of user queries if it does not lead to ambiguity, e.g., UUID names. 

\item{\textbf{Force VFM Thinking}} To ensure the successful trigger of VFMs for Visual ChatGPT, we append a suffix prompt to $(\mathcal{Q}_i)$:
 ``Since Visual ChatGPT is a text language model, Visual ChatGPT must use tools to observe images rather than imagination. The thoughts and observations are only visible for Visual ChatGPT, Visual ChatGPT should remember to repeat important information in the final response for Human. 
Thought: Do I need to use a tool?''. This prompt serves two purposes: 1) it prompts Visual ChatGPT to use foundation models instead of relying solely on its imagination; 2) it encourages Visual ChatGPT to provide specific outputs generated by the foundation models, rather than generic responses such as ``here you are''.

\end{itemize}

\subsection{Prompt Managing of Foundation Model Outputs $\mathcal{M}(\mathcal{F}(\mathcal{A}_{i}^{(j)}))$}
For the intermediate outputs from different VFMs $\mathcal{F}(\mathcal{A}_{i}^{(j)})$, Visual ChatGPT will implicitly summarize and feed them to the ChatGPT for subsequent interaction, i.e., calling other VFMs for further operations until reaching the ending condition or giving the feedback to the users. The inner steps can be summarized below:

\begin{itemize}
    \item \textbf{Generate Chained Filename} Since the intermediate outputs of Visual ChatGPT will become the inputs for the next implicit conversational round, we should make those outputs more logical to help the LLMs better understand the reasoning process. Specifically, the image generated from the Visual Foundation Models are saved under the``image/" folder, which hints the following strings representing an image name. Then, the image is named as ``\{Name\}\_\{Operation\}\_\{Prev\_Name\}\_\{Org\_Name\}", where \{Name\} is the UUID name mentioned above, with \{Operation\} as the operation name, \{Prev\_Name\} as the input image unique identifier, and \{Org\_Name\} as the original name of the image uploaded by users or generated by VFMs. For instance, ``image/ui3c\_edge-of\_o0ec\_nji9dcgf.png'' is a canny edge image named ``ui3c'' of input ``o0ec'', and the original name of this image is ``nji9dcgf''. With such a naming rule, it can hint ChatGPT of the intermediate result attributes,i.e., image, and how it was generated from a series of operations.
    
    \item \textbf{Call for More VFMs} One core of Visual ChatGPT is that it can automatically call for more VFMs to finish the user's command. More concretely, we make the ChatGPT keep asking itself whether it needs VFMs to solve the current problem by extending one suffix ``Thought: '' at the end of each generation.

    \item \textbf{Ask for More Details} When the user's command is ambiguous, Visual ChatGPT should ask the users for more details to help better leverage VFMs. This design is safe and critical since the LLMs are not permitted to arbitrarily tamper with or speculate about the user's intention without basis, especially when the input information is insufficient.
   
\end{itemize}

\section{Experiments}
\subsection{Setup}

We implement the LLM with ChatGPT~\cite{ouyang2022training} (OpenAI ``text-davinci-003" version), and guide the LLM with LangChain~\cite{Chase_LangChain_2022}\footnote{\url{https://github.com/hwchase17/langchain}}. We collect foundation models from HuggingFace Transformers~\cite{wolf-etal-2020-transformers}\footnote{\url{https://huggingface.co/docs/diffusers/index}}, Maskformer~\cite{cheng2021per}\footnote{\url{https://github.com/facebookresearch/MaskFormer}} and ControlNet~\cite{zhang2023adding}\footnote{\url{https://github.com/lllyasviel/ControlNet}}. The fully deployment of all the 22 VFMs requires 4 Nvidia V100 GPUs, but users are allowed to deploy fewer foundation models to save GPU resources flexibly. The maximum length of chat history is 2,000 and excessive tokens are truncated to meet the input length of ChatGPT.

\subsection{A full Case of Multiple Rounds Dialogue}
\begin{figure*}
    \centering
\includegraphics[width=1.0\textwidth]{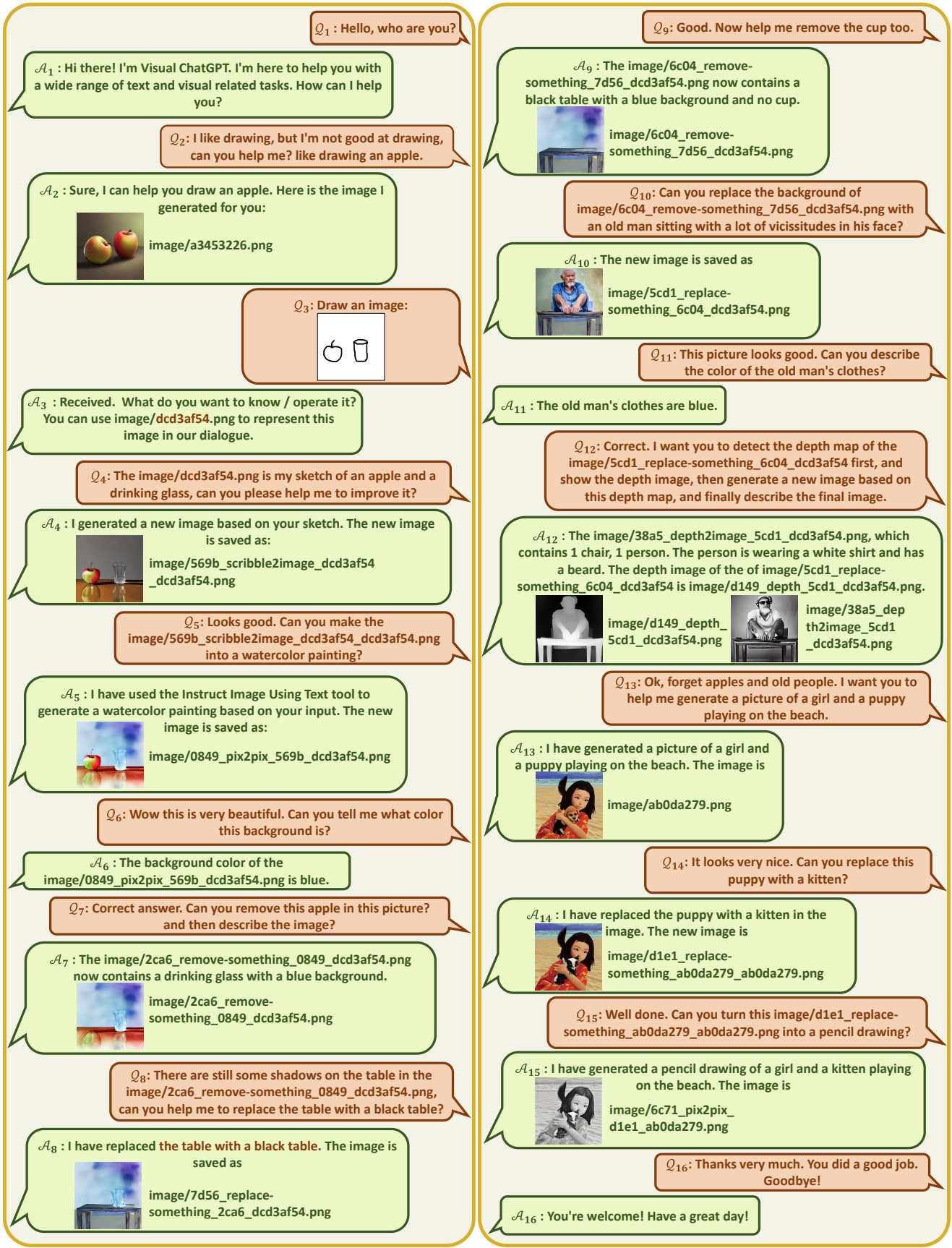} 
\caption{Multiple rounds of dialogue between humans and Visual ChatGPT. In the dialogues, Visual ChatGPT can understand human intents, support the language and image inputs, and accomplish complex visual tasks such as generation, question, and editing.} 
\label{fig:demo_poster_design}
\end{figure*}

Fig.~\ref{fig:demo_poster_design} shows a 16-round multi-modal dialogue case of Visual ChatGPT. In this case, the user asks for both text and image questions, and Visual ChatGPT responses with both text and images. The dialogue involves a discussion of multiple images, processing with multiple fundamental models, and handling questions that require multi-steps.

\begin{figure*}
    \centering
\includegraphics[width=1.0\textwidth]{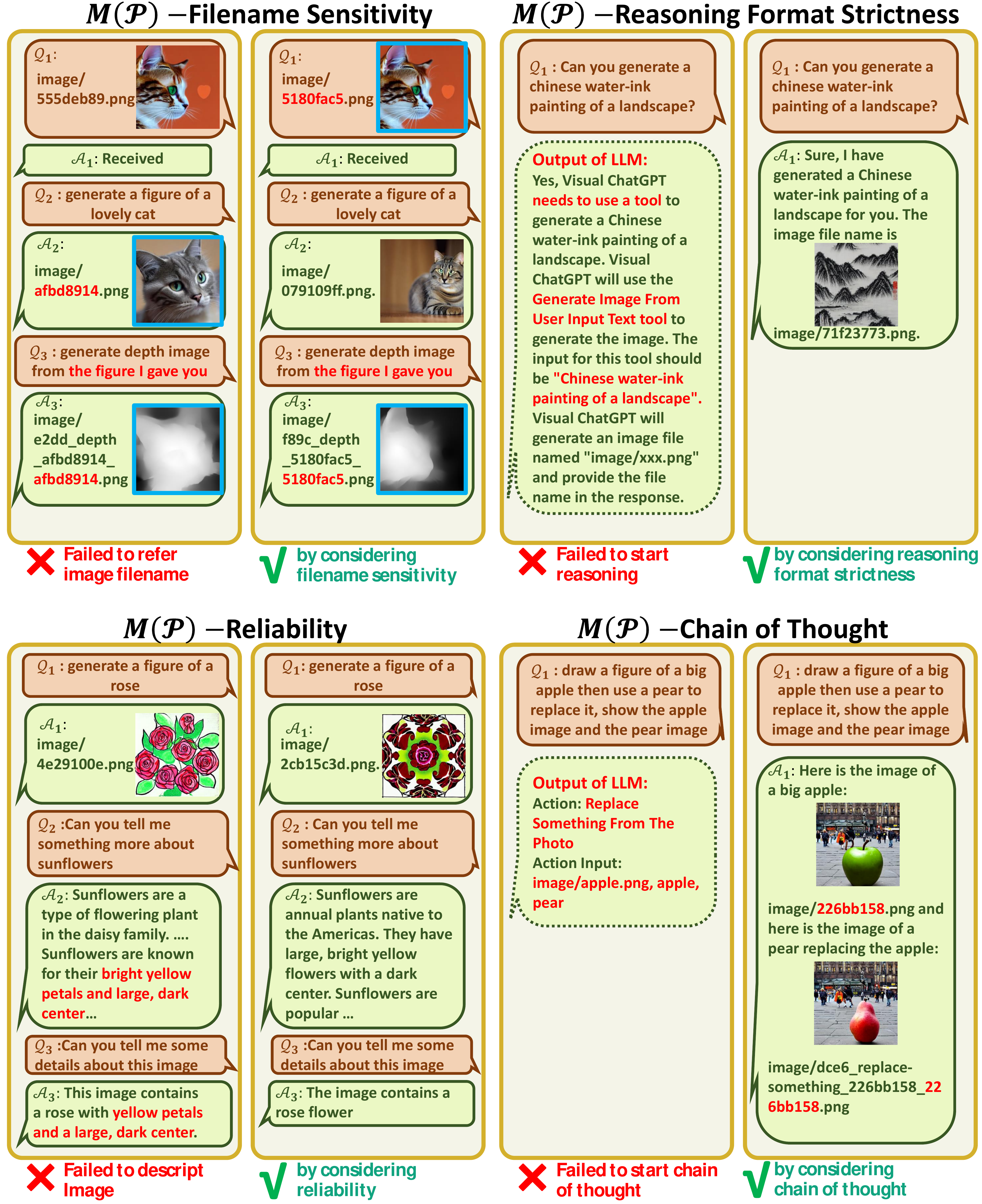} 
\caption{Case study of prompt managing of system principles. We qualitatively analyze the four proposals: file name sensitivity, reasoning format strictness, reliability, and chain of thoughts. The top-left shows whether emphasizing the file name sensitivity in $\mathcal{M}(\mathcal{P})$ affects the file reference accuracy. Further parsing cannot be performed for the top-right points without reasoning format strictness. The bottom-left shows the difference in whether to tell Visual ChatGPT to be loyal to tool observation rather than faking image content. The bottom-right shows emphasizing the ability to use tools in a chain will help the decision. } 
\label{fig:m(p)}
\end{figure*}

\begin{figure*}
    \centering
\includegraphics[width=1.0\textwidth]{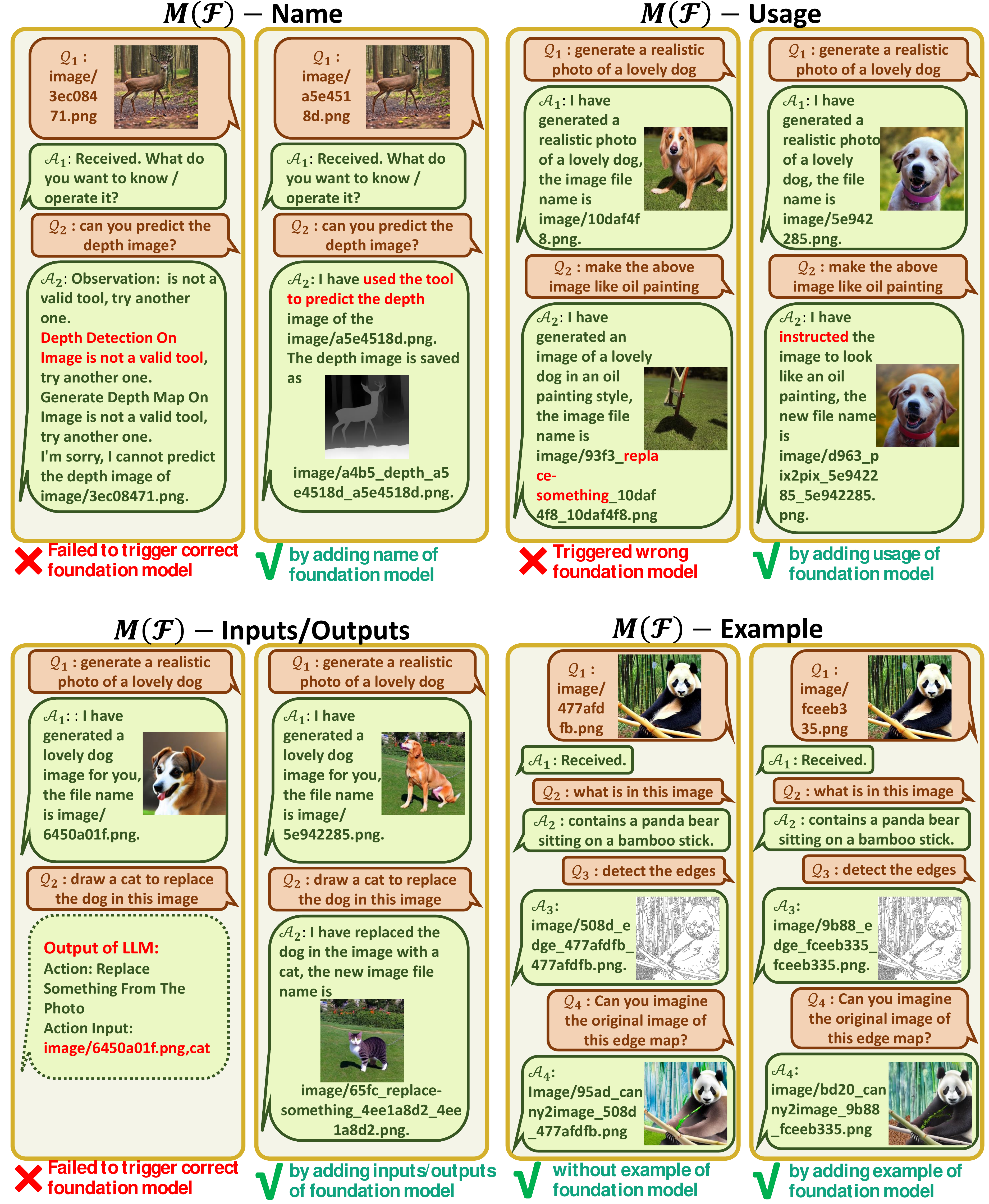} 
\caption{Case study of prompt managing of foundation models. We qualitatively analyze the four proposals: Name, Usage, Inputs/Outputs, and Example. The top-left shows that Visual ChatGPT will guess the tool name and then fails to use the correct tool without the tool name. The top-right shows that when the usage of the tool name is missing or unclear, it will call other tools or encounters an error. The bottom-left shows that the lack of inputs/outputs format requirements will lead to wrong parameters. The bottom-right shows that the example sometimes is optional because the ChatGPT is able to summarize the historical information and human intents to use the correct tool.} 
\label{fig:m(f)}
\end{figure*}

\begin{figure*}
    \centering
\includegraphics[width=1.0\textwidth]{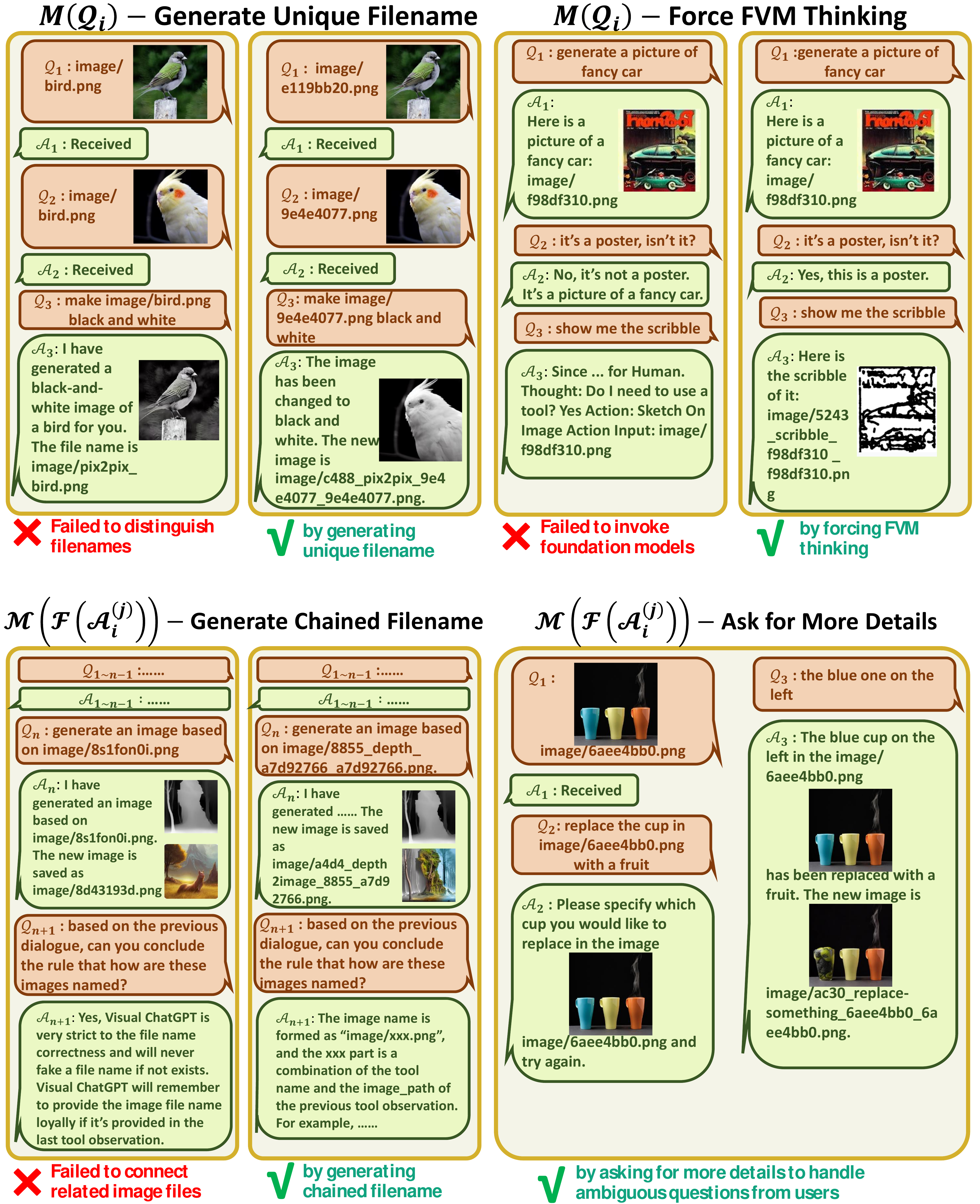} 
\caption{Case study of prompt managing of the user query and model output. We qualitatively analyze the four proposals: unique file names, force VFM thinking, chained file names, and ask for more details. The top-left shows that unique file names avoid overwriting. The top-right shows force VFM thinking encourages tool invoking and strict thinking format. The bottom-left shows chained file naming helps understand files and can be successfully observed and concluded by Visual ChatGPT. The bottom-right shows Visual ChatGPT is able to detect ambiguous references and ask for more details. } 
\label{fig:m(query_model_output)}
\end{figure*}

\subsection{Case Study of Prompt Manager}

\textbf{Case Study of prompt managing of system principles} is analyzed in Fig.~\ref{fig:m(p)}. To validate the effectiveness of our system principle prompts, we remove different parts from it to compare model performance. Each removal will result in different capacity degradation. 

\textbf{Case Study of prompt managing of foundation models} 
 is analyzed in Fig.~\ref{fig:m(f)}. The name of the VFM is the most important and needs to be clearly defined. When the name is missing or ambiguous, Visual ChatGPT will guess it many times until it finds an existing VFM, or encounters an error, as the top-left figure shows. The VFM usage should clearly describe the specific scenario where a model should be used to avoid the wrong responses. The top-right figure shows that the style transfer is mishandled to the replacement. The input and output format should be prompted accurately to avoid parameter errors, as shown in the bottom-left. The example prompt can help the model deal with complex usages but is optional. As shown in the bottom-right figure, although we delete the example prompt, ChatGPT can also summarize the dialogue history and human intents to use the correct VFM. The complete visual foundation model prompts is shown in Appendix~\ref{appdex.prompt}.    

\textbf{Case Study of prompt managing of user query} is analyzed in Fig\ref{fig:m(query_model_output)} upper part. The top-left figure shows that without image file unique naming, newly uploaded image file might be renamed to avoid overwritten and result in wrong reference. As shown in the top-right figure, by moving the thought guidance from $\mathcal{M}(\mathcal{P})$ to $\mathcal{M}(\mathcal{Q})$ and making it spoken in Visual ChatGPT's voice as a force thinking,  invoking more VFM is emphasized rather than imagination based on textual context as compared in $\mathcal{Q}_2$. By forcing Visual ChatGPT to say ``Thought: Do I need to use a tool?'',  $\mathcal{M}(\mathcal{Q})$ makes it easier to pass regex match correctly. In contrast, without force thinking, $\mathcal{A}_3$ may wrongly generate the end of thoughts token and directly consider all of its ChatGPT outputs as the final response.

\textbf{Case Study of prompt managing of model outputs} is analyzed in Fig\ref{fig:m(query_model_output)} bottom part. The bottom-left picture compares the performance of removing and keeping the chained naming rule. With the chained naming rule, Visual ChatGPT can recognize the file type, trigger the correct VFM, and conclude the file dependency relationship naming rule. It shows that the chained naming rule does help Visual ChatGPT to understand. The bottom-right picture gives an example of asking for more details when the item inference is ambiguous, which also indicates the safety of our system.

\section{Limitations}

Although Visual ChatGTP is a promising approach for multi-modal dialogue, it has some limitations, including:

\begin{itemize}
\item{\textbf{Dependence on ChatGPT and VFMs}} Visual ChatGPT relies heavily on ChatGPT to assign tasks and on VFMs to execute them. The performance of Visual ChatGPT is thus heavily influenced by the accuracy and effectiveness of these models.

\item{\textbf{Heavy Prompt Engineering}} Visual ChatGPT requires a significant amount of prompt engineering to convert VFMs into language and make these model descriptions distinguishable. This process can be time-consuming and requires expertise in both computer vision and natural language processing.

\item{\textbf{Limited Real-time Capabilities}} Visual ChatGPT is designed to be general. It tries to decompose a complex task into several subtasks automatically. Thus, when handling a specific task, Visual ChatGPT may invoke multiple VFMs, resulting in limited real-time capabilities compared to expert models specifically trained for a particular task.

\item{\textbf{Token Length Limitation}} The maximum token length in ChatGPT may limit the number of foundation models that can be used. If there are thousands or millions of foundation models, a pre-filter module may be necessary to limit the VFMs fed to ChatGPT.

\item{\textbf{Security and Privacy}} The ability to easily plug and unplug foundation models may raise security and privacy concerns, particularly for remote models accessed via APIs. Careful consideration and automatic check must be given to ensure that sensitive data should not be exposed or compromised.

\end{itemize}






\section{Conclusion}

In this work, we propose Visual ChatGPT, an open system incorporating different VFMs and enabling users to interact with ChatGPT beyond language format. To build such a system, we meticulously design a series of prompts to help inject the visual information into ChatGPT, which thus can solve the complex visual questions step-by-step. Massive experiments and selected cases have demonstrated the great potential and competence of Visual ChatGPT for different tasks.
Apart from the aforementioned limitations, another concern is that some generation results are unsatisfied due to the failure of VFMs and the instability of the prompt. Thus, one self-correction module is necessary for checking the consistency between execution results and human intentions and accordingly making the corresponding editing. Such self-correction behavior can lead to more complex thinking of the model, significantly increasing the inference time. We will solve such an issue in the future.

{\small
\bibliographystyle{ieee_fullname}
\bibliography{egbib}
}

\clearpage
\appendix

\section{Tool Details}~\label{appdex.prompt}

\begin{description}
      \item[$\bullet$ Remove Something From The Photo:]\ 
    \begin{itemize}
      \item[\textbf{Model:}]  ``runwayml/stable-diffusion-inpainting'' from Huggingface library, StableDiffusionInpaintPipeline model; ``CIDAS/clipseg-rd64-refined'' from Huggingface library, CLIPSegForImageSegmentation model.
      \item[\textbf{InOut:}] image\_path, textual what to remove $\rightarrow$ image\_path
      \item[\textbf{Prompt:}]  Remove something from the photo: useful for when you want to remove and object or something from the photo from its description or location. The input to this tool should be a comma seperated string of two, representing the image\_path and the object need to be removed.
    \end{itemize}
      \item[$\bullet$ Replace Something From The Photo:]\ 
    \begin{itemize}
      \item[\textbf{Model:}]  "runwayml/stable-diffusion-inpainting" from Huggingface library, StableDiffusionInpaintPipeline model; ``CIDAS/clipseg-rd64-refined'' from Huggingface library, CLIPSegForImageSegmentation model.
      \item[\textbf{InOut:}] image\_path, textual what to replace, textual what to add $\rightarrow$ image\_path
      \item[\textbf{Prompt:}] Replace something from the photo: useful for when you want to replace an object from the object description or location with another object from its description. The input to this tool should be a comma seperated string of three, representing the image\_path, the object to be replaced, the object to be replaced with.
    \end{itemize}
      \item[$\bullet$ Instruct Image Using Text:]\ 
    \begin{itemize}
      \item[\textbf{Model:}]  "timbrooks/instruct-pix2pix" from HuggingFace, StableDiffusionInstructPix2PixPipeline model.
      \item[\textbf{InOut:}] image\_path, textual how to modify $\rightarrow$ image\_path
      \item[\textbf{Prompt:}] Instruct image using text: useful for when you want to the style of the image to be like the text. like: make it look like a painting. or make it like a robot. The input to this tool should be a comma seperated string of two, representing the image\_path and the text.
    \end{itemize}
      \item[$\bullet$ Answer Question About The Image:]\ 
    \begin{itemize}
      \item[\textbf{Model:} ] "Salesforce/blip-vqa-base" from HuggingFace, BlipForQuestionAnswering model.  
      \item[\textbf{InOut:}] image\_path, question $\rightarrow$  answer
      \item[\textbf{Prompt:} ] useful when you need an answer for a question based on an image like: what is the background color of the last image, how many cats in this figure, what is in this figure. 
    \end{itemize}
    \item[$\bullet$ Get Photo Description:]\ 
    \begin{itemize}
      \item[\textbf{Model:}]  "Salesforce/blip-image-captioning-base" from HuggingFace library, BlipForConditionalGeneration model. 
      \item[\textbf{InOut:}] image\_path 	$\rightarrow$ natural language description
      \item[\textbf{Prompt:}] Get photo description: useful for when you want to know what is inside the photo. The input to this tool should be a string, representing the image\_path.
    \end{itemize}
  \item[$\bullet$ Generate Image From User Input Text:]\ 
    \begin{itemize}
      \item[\textbf{Model:}]  "runwayml/stable-diffusion-v1-5" from HuggingFace library, StableDiffusionPipeline model.
      \item[\textbf{InOut:}] textual description 	$\rightarrow$ image path 
      \item[\textbf{Prompt:}] Generate image from user input text: useful for when you want to generate an image from a user input text and it saved it to a file. The input to this tool should be a string, representing the text used to generate image.
    \end{itemize}
    
    \item[$\bullet$ Edge Detection On Image :]\ 
    \begin{itemize}
      \item[\textbf{Model:} ] Canny Edge Detector from OpenCV
      \item[\textbf{InOut:}]  image\_path $\rightarrow$ edge\_image\_path
      \item[\textbf{Prompt:} ]  Edge Detection On Image : useful for when you want to detect the edge of the image. like: detect the edges of this image, or canny detection on image, or peform edge detection on this image, or detect the canny image of this image. The input to this tool should be a string, representing the image\_path.
    \end{itemize}
    
    \item[$\bullet$ Image Generation Condition On Canny Image:]\ 
    \begin{itemize}
      \item[\textbf{Model:}] ControlNet for Canny Edge.
      \item[\textbf{InOut:}] edge\_image\_path, textual description $\rightarrow$ image\_path
      \item[\textbf{Prompt:}] useful for when you want to generate a new real image from both the user desciption and a canny image. like: generate a real image of a object or something from this canny image, or generate a new real image of a object or something from this edge image. The input to this tool should be a comma seperated string of two, representing the image\_path and the user description.
    \end{itemize}

    \item[$\bullet$ Line Detection On Image :]\ 
    \begin{itemize}
      \item[\textbf{Model:}] M-LSD Detector for Straight Line
      \item[\textbf{InOut:}]  image\_path $\rightarrow$ line\_image\_path
      \item[\textbf{Prompt:} ]  Line Detection On Image : useful for when you want to detect the straight line of the image. like: detect the straight lines of this image, or straight line detection on image, or peform straight line detection on this image, or detect the straight line image of this image. The input to this tool should be a string, representing the image\_path
    \end{itemize}
    
    \item[$\bullet$ Generate Image Condition On Line Image:]\ 
    \begin{itemize}
      \item[\textbf{Model:}] ControlNet for M-LSD Lines.
      \item[\textbf{InOut:}] line\_image\_path, textual description $\rightarrow$ image\_path
      \item[\textbf{Prompt:}] useful for when you want to generate a new real image from both the user desciption and a straight line image. like: generate a real image of a object or something from this straight line image, or generate a new real image of a object or something from this straight lines. The input to this tool should be a comma seperated string of two, representing the image\_path and the user description. 
    \end{itemize}

    \item[$\bullet$ Hed Detection On Image :]\ 
    \begin{itemize}
      \item[\textbf{Model:}] HED Boundary Detector
      \item[\textbf{InOut:}]  image\_path $\rightarrow$ hed\_image\_path
      \item[\textbf{Prompt:}]   Hed Detection On Image: useful for when you want to detect the soft hed boundary of the image. like: detect the soft hed boundary of this image, or hed boundary detection on image, or peform hed boundary detection on this image, or detect soft hed boundary image of this image. The input to this tool should be a string, representing the image\_path
    \end{itemize}

    \item[$\bullet$ \footnotesize{Generate Image Condition On Soft Hed Boundary Image} :]\ 
    \begin{itemize}
      \item[\textbf{Model:}] ControlNet for HED.
      \item[\textbf{InOut:}]  hed\_image\_path, textual description $\rightarrow$ image\_path
      \item[\textbf{Prompt:} ] Generate Image Condition On Soft Hed Boundary Image: useful for when you want to generate a new real image from both the user desciption and a soft hed boundary image. like: generate a real image of a object or something from this soft hed boundary image, or generate a new real image of a object or something from this hed boundary.  The input to this tool should be a comma seperated string of two, representing the image\_path and the user description
    \end{itemize}

    \item[$\bullet$ Segmentation On Image :]\ 
    \begin{itemize}
      \item[\textbf{Model:}] Uniformer Segmentation 
      \item[\textbf{InOut:}]  image\_path $\rightarrow$ segment\_image\_path
      \item[\textbf{Prompt:}]  useful for when you want to detect segmentations of the image. like: segment this image, or generate segmentations on this image, or peform segmentation on this image. The input to this tool should be a string, representing the image\_path
    \end{itemize}

    \item[$\bullet$ Generate Image Condition On Segmentations :]\ 
    \begin{itemize}
          \item[\textbf{Model:}] ControlNet for Segmentation.
      \item[\textbf{InOut:}]  segment\_image\_path, textual description $\rightarrow$ image\_path
      \item[\textbf{Prompt:}]  useful for when you want to generate a new real image from both the user desciption and segmentations. like: generate a real image of a object or something from this segmentation image, or generate a new real image of a object or something from these segmentations. The input to this tool should be a comma seperated string of two, representing the image\_path and the user description
    \end{itemize}

    \item[$\bullet$ Predict Depth On Image :]\ 
    \begin{itemize}
      \item[\textbf{Model:}] MiDaS Depth Estimation
      \item[\textbf{InOut:}]  image\_path $\rightarrow$ depth\_image\_path
      \item[\textbf{Prompt:}]  Predict Depth Map On Image : useful for when you want to detect depth of the image. like: generate the depth from this image, or detect the depth map on this image, or predict the depth for this image, the input to this tool should be a string, representing the image\_path.
    \end{itemize}

    \item[$\bullet$ Generate Image Condition On Depth:]\ 
    \begin{itemize}
      \item[\textbf{Model:}] ControlNet for Depth.
      \item[\textbf{InOut:}]  depth\_image\_path, textual description $\rightarrow$ image\_path
      \item[\textbf{Prompt:}]  Generate Image Condition On Depth Map : useful for when you want to generate a new real image from both the user desciption and depth image. like: generate a real image of a object or something from this depth image, or generate a new real image of a object or something from the depth map, The input to this tool should be a comma seperated string of two, representing the image\_path and the user description.
    \end{itemize}
    
    \item[$\bullet$ Predict Normal Map On Image :]\ 
    \begin{itemize}
      \item[\textbf{Model:}] MiDaS Depth Estimation for Normal Map
      \item[\textbf{InOut:}]  image\_path $\rightarrow$ norm\_image\_path
      \item[\textbf{Prompt:}]  Predict Normal Map On Image : useful for when you want to detect norm map of the image. like: generate normal map from this image, or predict normal map of this image The input to this tool should be a string, representing the image\_path
    \end{itemize}

    \item[$\bullet$ Generate Image Condition On Normal Map :]\ 
    \begin{itemize}
      \item[\textbf{Model:}] ControlNet for Normal Map.
      \item[\textbf{InOut:}]  norm\_image\_path, textual description $\rightarrow$ image\_path
      \item[\textbf{Prompt:}]  Generate Image Condition On Normal Map : useful for when you want to generate a new real image from both the user desciption and normal map. like: generate a real image of a object or something from this normal map, or generate a new real image of a object or something from the normal map. The input to this tool should be a comma seperated string of two, representing the image\_path and the user description
    \end{itemize}

     \item[$\bullet$ Sketch Detection On Image :]\ 
    \begin{itemize}
      \item[\textbf{Model:}] HED Boundary Detector
      \item[\textbf{InOut:}]  image\_path $\rightarrow$ sketch\_image\_path
      \item[\textbf{Prompt:}]  Sketch Detection On Image: useful for when you want to generate a scribble of the image. like: generate a scribble of this image, or generate a sketch from this image, detect the sketch from this image. The input to this tool should be a string, representing the image\_path
    \end{itemize}

     \item[$\bullet$ Generate Image Condition On Sketch Image :]\ 
    \begin{itemize}
      \item[\textbf{Model:}] ControlNet for Scribble.
      \item[\textbf{InOut:}]  sketch\_image\_path, textual description $\rightarrow$ image\_path
      \item[\textbf{Prompt:}]  useful for when you want to generate a new real image from both the user desciption and a scribble image. like: generate a real image of a object or something from this scribble image, or generate a new real image of a object or something from this sketch. The input to this tool should be a comma seperated string of two, representing the image\_path and the user description
    \end{itemize}

     \item[$\bullet$ Pose Detection On Image :]\ 
    \begin{itemize}
      \item[\textbf{Model:}] Openpose Detector
      \item[\textbf{InOut:}]  image\_path $\rightarrow$ pos\_image\_path
      \item[\textbf{Prompt:}]  Pose Detection On Image: useful for when you want to detect the human pose of the image. like: generate human poses of this image, or generate a pose image from this image. The input to this tool should be a string, representing the image\_path
    \end{itemize}

     \item[$\bullet$ Generate Image Condition On Pose Image :]\ 
    \begin{itemize}
      \item[\textbf{Model:}] ControlNet for Human Pose.
      \item[\textbf{InOut:}]  pos\_image\_path, textual description $\rightarrow$ image\_path
      \item[\textbf{Prompt:}]  Generate Image Condition On Pose Image: useful for when you want to generate a new real image from both the user desciption and a human pose image. like: generate a real image of a human from this human pose image, or generate a new real image of a human from this pose. The input to this tool should be a comma seperated string of two, representing the image\_path and the user description
    \end{itemize}

\end{description}

\end{document}